\documentclass[a4paper,twoside]{article}
\usepackage{epsfig}
\usepackage{subcaption}
\usepackage{calc}
\usepackage{amssymb}
\usepackage{amstext}
\usepackage{amsmath}
\usepackage{amsthm}
\usepackage{multicol}
\usepackage{pslatex}
\usepackage{apalike}
\usepackage{adjustbox}
\usepackage{url}
\usepackage{xspace}
\usepackage{color}
\usepackage{SCITEPRESS}     % Please add other packages that you may need BEFORE the SCITEPRESS.sty package.
\makeatletter
\DeclareRobustCommand\onedot{\futurelet\@let@token\@onedot}
\def\@onedot{\ifx\@let@token.\else.\null\fi\xspace}

\def\ie{\emph{i.e}\onedot}

\def\etc{\emph{etc}\onedot}

\makeatother

\begin{document}
\title{STaDA: Style Transfer as Data Augmentation}

\author{\authorname{Xu Zheng\sup{1,2}, Tejo Chalasani\sup{2}, Koustav Ghosal\sup{2}, Sebastian Lutz\sup{2}, Aljosa Smolic\sup{2}}
\affiliation{\sup{1}Accenture Labs, Dublin, Ireland}
\affiliation{\sup{2}School of Computer Science and Statistics, Trinity College Dublin, Ireland}
\email{xu.b.zheng@accenture.com, \{chalasat, ghosalk, lutzs, smolica\}@scss.tcd.ie}
}


\abstract{The success of training deep Convolutional Neural Networks (CNNs) heavily depends on a significant amount of labelled data. Recent research has found that neural style transfer algorithms can apply the artistic style of one image to another image without changing the latter's high-level semantic content, which makes it feasible to employ neural style transfer as a data augmentation method to add more variation to the training dataset. The contribution of this paper is a thorough evaluation of the effectiveness of the neural style transfer as a data augmentation method for image classification tasks. We explore the state-of-the-art neural style transfer algorithms and apply them as a data augmentation method on Caltech 101 and Caltech 256 dataset, where we found around 2\% improvement from 83\% to 85\% of the image classification accuracy with VGG16, compared with traditional data augmentation strategies. We also combine this new method with conventional data augmentation approaches to further improve the performance of image classification. This work shows the potential of neural style transfer in computer vision field, such as helping us to reduce the difficulty of collecting sufficient labelled data and improve the performance of generic image-based deep learning algorithms.}

\onecolumn \maketitle \normalsize \vfill

\section{\uppercase{Introduction}}
\label{sec:introduction}

\noindent Data augmentation refers to the task of adding diversity to the training data of a neural network, especially when there is a paucity of sufficient samples. Popular deep architectures such as AlexNet \cite{krizhevsky2012imagenet} or VGGNet \cite{simonyan2014very} have millions of parameters and thus require a reasonably large dataset to be trained for a particular task. Lack of adequate data leads to overfitting \ie high training accuracy but poor generalisation over the test samples \cite{caruana2001overfitting}. In many computer vision tasks, gathering raw data can be very time-consuming and expensive. For example, in the domain of medical image analysis, in critical tasks such as cancer detection \cite{kyprianidis2013state} and cancer classification \cite{vasconcelos2017increasing}, researchers are often restricted by the lack of reliable data. Thus, it is a common practice to use data augmentation techniques such as flipping, rotation, cropping \etc to increase the variety of samples fed to the network. Recently, more complex techniques using a green screen with different random backgrounds to increase the number of training images has been introduced by \cite{chalasani2018Egocentric}. 

%\noindent Many image data augmentation techniques have been widely used in the state-of-the-art Convolutional Neural Network (CNN) architectures  . These traditional data augmentation techniques, such as flipping, rotation, and cropping, are found effective to generate additional labelled images and avoid over-fitting for a complex and deep CNN model. But these techniques can only add limited diversity to the original image distribution. With the fast development of deep CNN architectures, more effective data augmentation techniques are likely to be needed for future research. 

%For many images and video machine learning projects, gathering raw data can be very time-consuming and challenging. Like some critical image-based tasks like cancer detection \cite{kyprianidis2013state} and cancer classification \cite{vasconcelos2017increasing}, researchers are often restricted by the lack of reliable data. What is more, for many machine learning algorithms, if we only use the raw data, the model is easy to get over-fitting and does not have enough generalisation on unseen data. We realise that it is of great importance for people to access a significant amount of reliable and labelled image data, hence exploring more effective data augmentation techniques would be a valuable research and can help us to find potential solutions to address the problems mentioned above.

\begin{figure}
    \centering
    \def\arraystretch{1.4}
    \begin{tabular}{cc}
    \includegraphics[width = .20\textwidth, height = .15\textwidth]{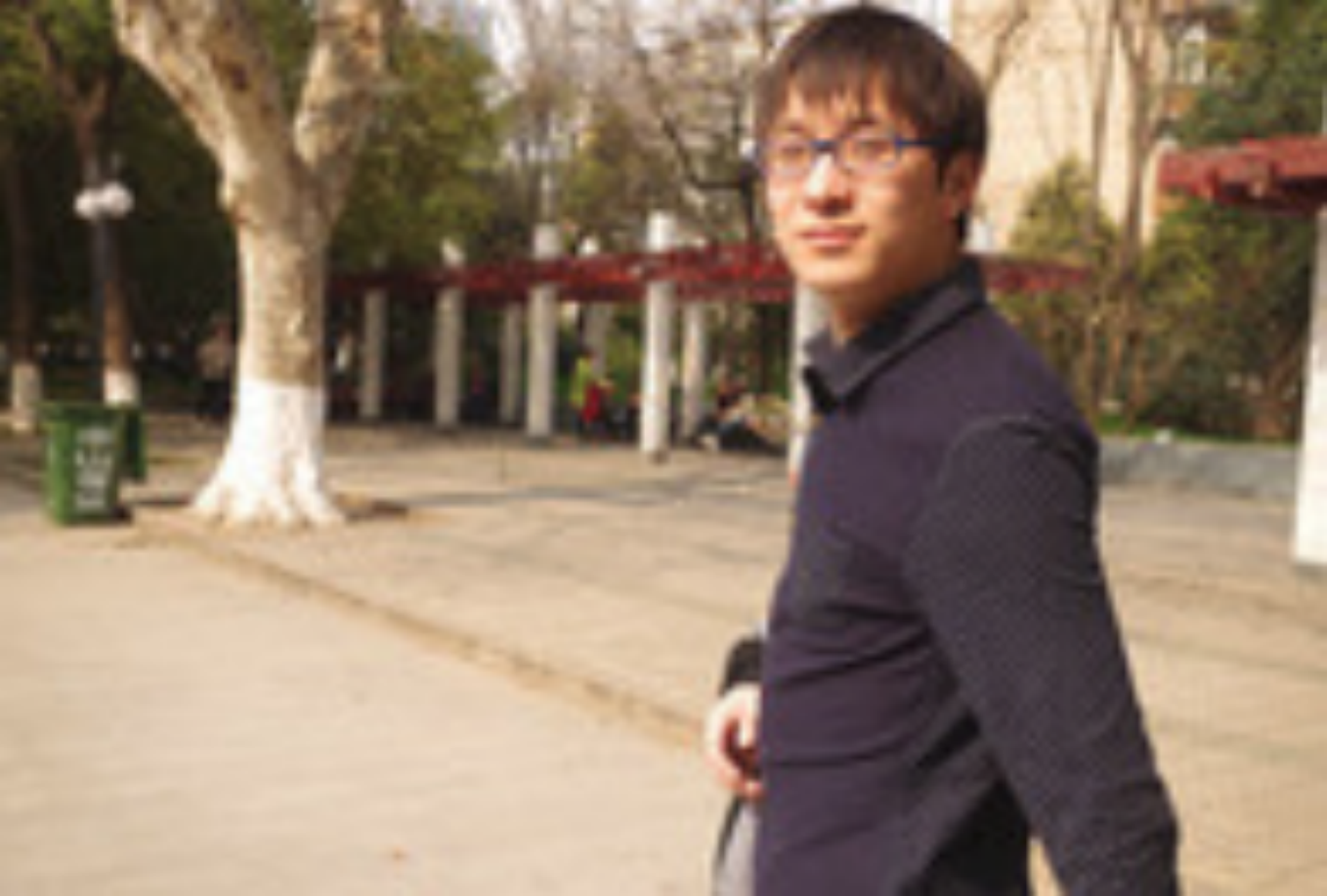}&
    \includegraphics[width = .20\textwidth, height = .15\textwidth]{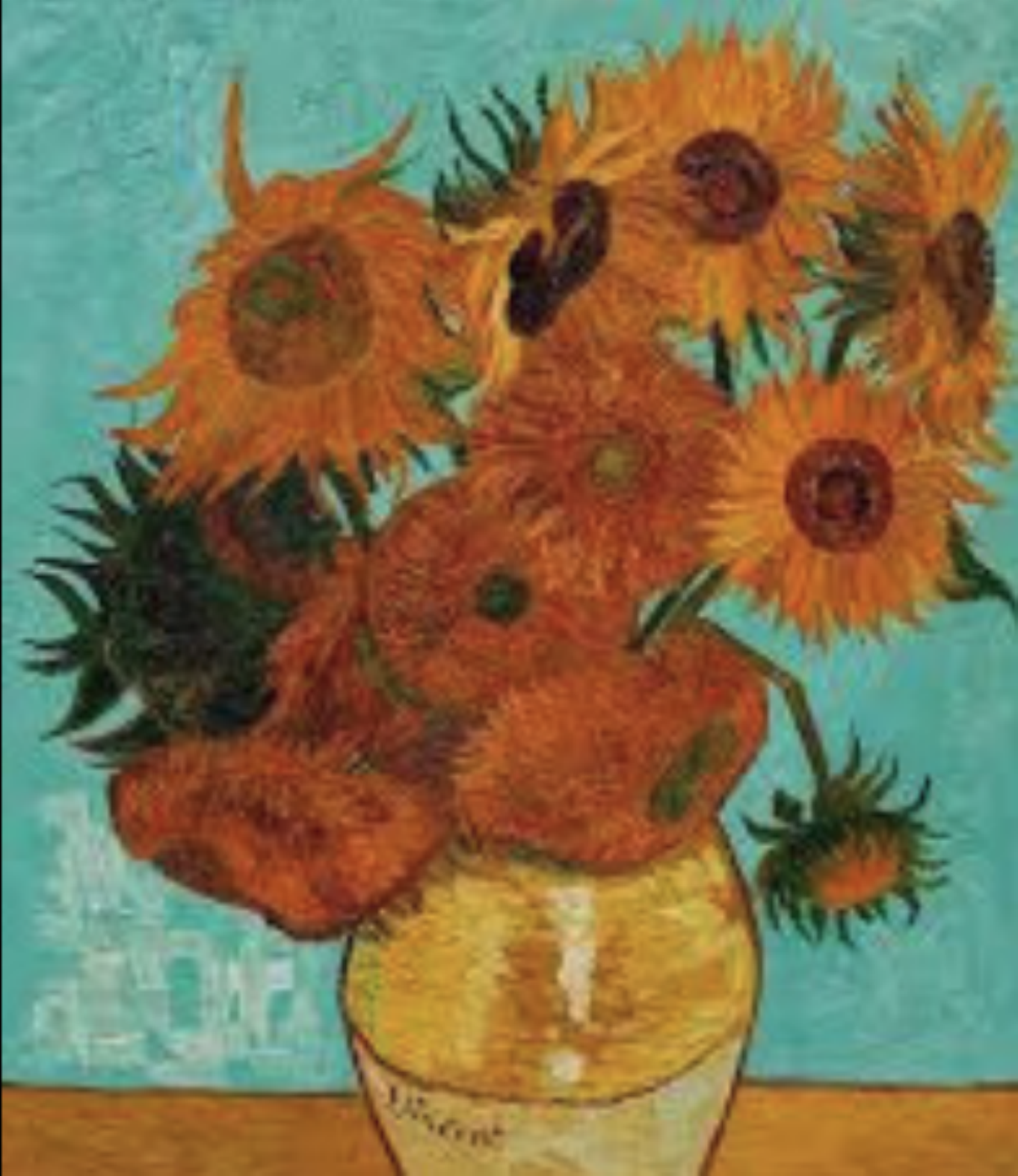}\\
    (a) Raw image  & (b) Style Image\\
    \multicolumn{2}{c}{\includegraphics[width = 0.40 \textwidth]{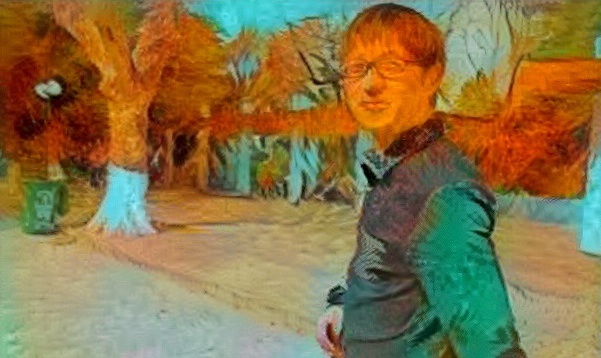}} \\
    \multicolumn{2}{c}{(c) Stylized Image}
    \end{tabular}
    \caption{Style Transfer: The style of Sunflower (b) (Van Gogh) is applied on the photo (a) to generate a new stylised image (c), which can keep the main content of the photo but as well as contain the texture style from image Sunflower.}
   \label{1.3}
\end{figure}

In this paper, we explore the capacity of neural style transfer \cite{gatys2016image} as an alternative data augmentation strategy. We propose a pipeline that applies this strategy on image classification tasks and verify its effectiveness on multiple datasets. Style transfer refers to the task of applying the artistic style of one image to another, without changing the high-level semantic content (Figure \ref{1.3}). The main idea of this algorithm is to jointly minimise the distance of the content representation and the style representation learned on different layers of a convolutional neural network, which allows translation from noise to the target image in a single pass through a network that is trained per style. 

The crux of this work is to use a style transfer network as a generative model to create more samples for training a CNN. Since style transfer preserves the overall semantic content of the original image, the high-level discriminative features of an object are maintained. On the other hand, by changing the artistic style of some randomly selected training samples it is possible to train a classifier which is invariant to undesirable components of the data distribution. For example, let us assume a scenario where a dataset of simple objects such as Caltech 256 \cite{griffin2007caltech} has a category \textit{car} but there are more images of \textit{red} cars than any other colour. The model trained on such a dataset will associate red colour to the car category, which is undesired for a generic car classifier. Using style transfer as the data augmentation method can be an effective strategy to avoid such associations. 

% (Figure \kg{refer to the appendix images. Also a suggestion , you can put the appendix figure in a 8 $\times$ 1 tabular within a figure environment})
In this work, we use eight different style images as palette to augment the original image datasets (Section \ref{ssec:datasets}). Additionally, we investigate if different image styles have different effects on image classification. This paper is organised as follows. In Section \ref{sec:related_work}, we discuss research related to our work. In Section \ref{sec:des_n_imp}, we describe different components of the architecture used, in Section \ref{sec:experiments} we describe our experiments and report and analyse the results. 

% \begin{figure}[!ht]
% \begin{subfigure}{.229\textwidth}
%   \centering
%   \includegraphics[width=.80\linewidth]{section1/rawImage.jpg}
%   \caption{Raw Images}
%   \label{1.1}
% \end{subfigure}%
% \begin{subfigure}{.229\textwidth}
%   \centering
%   \includegraphics[width=.80\linewidth]{section1/stylizedImage.jpg}
%   \caption{Stylised Image}
%   \label{1.2}
% \end{subfigure}
% \caption{Style Transfer: The style of Sunflower (Van Gogh) is applied on the photo and generate a new stylised image (b), which can keep the main content of the photo but as well as contain the texture style from image Sunflower.}
% \label{1.3}
% \end{figure}

\section{\uppercase{Related Work}}
\label{sec:related_work}

\begin{figure*}[ht]
  \centering
  \includegraphics[scale=0.56]{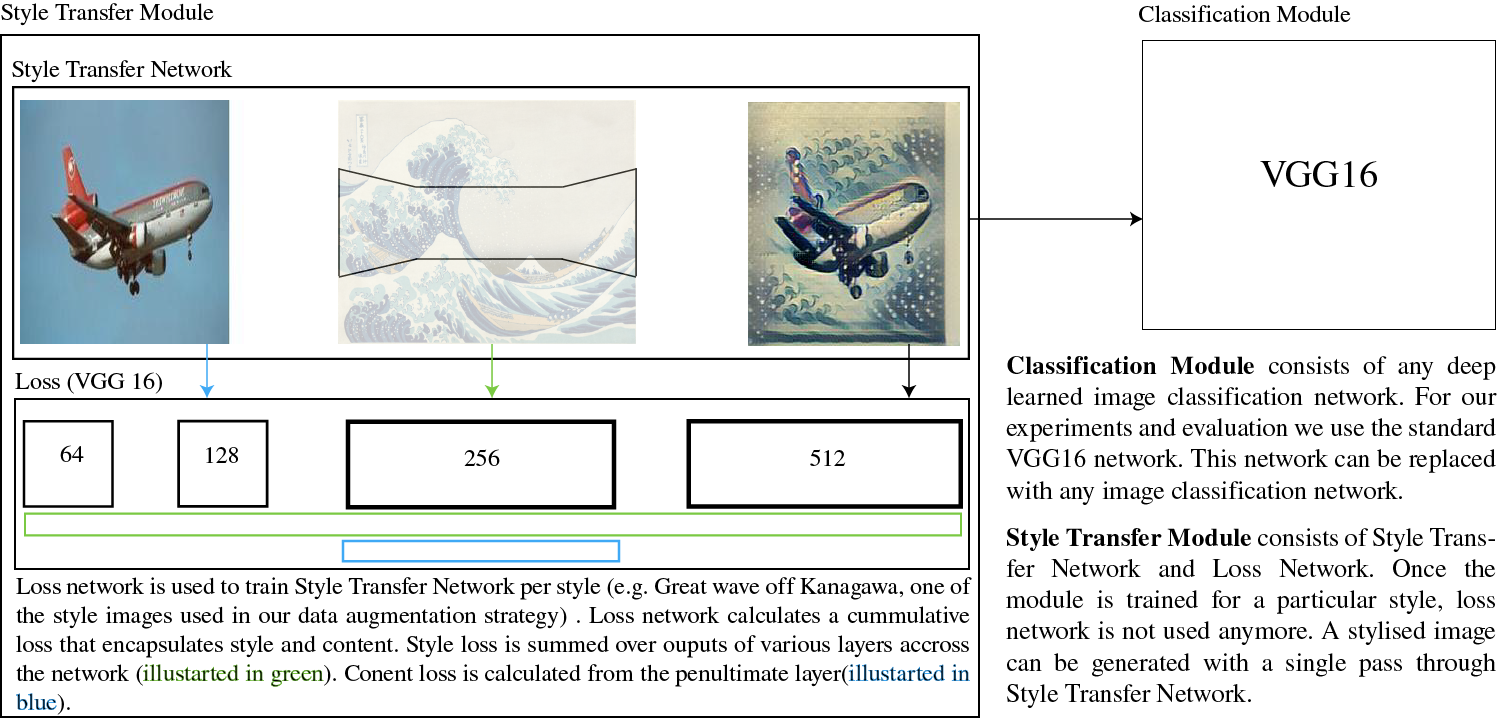}
  \caption{Components of the modular design for data augmentation using style transfer. The first module is a style transfer module that is trained per style and takes an image as input to generate a stylised image and the second module is a classification network that uses the stylised image for training to improve accuracy.}
  \label{3.1}
\end{figure*} 

\noindent In this section, we review the state-of-the-art research relevant to the problem addressed in this paper. In Section \ref{sec:related_work_style}, we review the development of neural style transfer algorithms and in Section \ref{sec:related_work_data_aug}, we discuss the traditional data augmentation strategies and how they affect the performance of a CNN in the standard computer vision tasks such as classification, detection, segmentation \etc. 
\subsection{Neural Style Transfer}
\label{sec:related_work_style}
% The main goal of style transfer is to apply the texture synthesis from one image onto another without changing the high-level content of the original image. During this process, we need to find a stylized image that contains the semantic content of a content image and texture style of an arbitrary image \cite{dumoulin2017learned}.
\noindent The main goal of style transfer is to synthesise a stylised image from two different images, one supplying the texture and another providing the high-level semantic content. Broadly, the state-of-the-art style transfer algorithms can be divided into two groups --- descriptive and generative. Descriptive approach refers to changing the pixels of a noise image in an iterative manner whereas generative approach achieves the same in a single forward pass by using a pre-trained model of the desired style \cite{JingYFYS17}.
% Many non-parametric algorithms can perform style transfer while preserving the structure of the content image, but the major fundamental limitation for them is that they only use low-level image features. However, the recent research has shown that the high-level information and texture synthesis of an image can be learned by CNNs optimised for computer visions tasks, which allows us to perform image style transfer with Convolutional Neural Networks (CNNs) \cite{krizhevsky2012imagenet} \cite{erhan2009visualizing} \cite{gatys2015texture}.

\textbf{Descriptive Approach:}
The work done by \cite{gatys2016image} is the first descriptive approach of neural style transfer. Starting from random noise, their algorithm transforms the random noise image in an iterative manner such that it mimics the content from one image and style or texture from another. The content is optimised by minimising the $\mathcal{L}_2$ distance between the high level CNN features of the content and stylised image. On the other hand, the style is optimised by matching the Gram matrices of the style and stylised image. Several algorithms followed directly from this approach by addressing some of its limitations. \cite{risser2017stable} propose a more stable optimisation strategy by adding histogram losses. \cite{li2017demystifying} closely investigate the different components of the work by \cite{gatys2016image} and present key insights on the optimisation methods, kernels used and normalisation strategies adopted.  \cite{yin2016content,chen2016towards} propose content-aware neural style transfer strategies which aim to preserve high-level semantic correspondence between the content and the target.

\textbf{Generative Approach:} 
The descriptive approach is limited by its slow computation time. In the generative or the \textit{faster}  approach, a model is trained in advance for each style image. While the inference in descriptive approach occurs slowly over several iterations, in this case, it is achieved through a single forward pass. \cite{johnson2016Perceptual} propose a two-component architecture --- generator and loss networks. They also introduce a novel loss function based on perceptual difference between the content and target. In \cite{ulyanov2016texture}, the authors improve the work by \cite{johnson2016Perceptual} by using a better generator network. \cite{li2016precomputed} train a Markovian feed-forward network by using an adversarial loss function. \cite{dumoulin2016learned} train for multiple styles using the same network.

% Based on the descriptive model, fast neural style transfer methods have been proposed recently. Instead of using image iteration method, these algorithms with higher efficiency are based on model iteration. In these works, the authors trained models that can generate stylized images directly so that we can get stylized images quickly. This method is called Generative Neural Style Transfer \cite{johnson2016Perceptual}. In a Generative Neural Style Transfer system, there are mainly two components: an image transformation network and a loss network. The image transformation network is usually a deep residual convolutional neural network parameterised by weights. It is used to transform input images into output images, which will be fed to the loss network. The loss function in the loss network computes a scalar value to measure the difference between the output image and a target image, including content reconstruction loss and style reconstruction loss.

\subsection{Data Augmentation}
\label{sec:related_work_data_aug}
\noindent As discussed previously in Section \ref{sec:introduction}, data augmentation refers to the process of adding more variation to the training data in order to improve the generalisation capabilities of the trained model. It is particularly useful in scenarios where there is a scarcity of training samples. Some common strategies for data augmentation are flipping, re-scaling, cropping, \etc. Research related to developing novel data augmentation strategies is almost as old as the research in \textit{deeper} neural networks with more parameters. 

In \cite{krizhevsky2012imagenet}, the authors applied two different augmentation methods to improve the performance of their model. The first one is horizontal flip, without which their network showed substantial overfitting even with only five layers. The second strategy is to perform a PCA on the RGB values of the image pixels in the training set, and use the top principal components, which reduced over $1\%$ of the top-$1$ error rate. Similarly, the ZFNet \cite{zeiler2014visualizing} and VGGNet \cite{simonyan2014very} also apply multiple different crops and flips for each training image to boost training set size and improve the performance of their models. These well-known CNNs architectures achieved outstanding results in the ImageNet challenge \cite{deng2009imagenet}, and their success demonstrated the effectiveness and importance of data augmentation.

Besides the traditional \textit{transformative} data augmentation strategies as discussed above, some recently proposed methods follow a \textit{generative} approach. In the work \cite{perez2017effectiveness}, the authors explore Generative Adversarial Nets to generate stylised images to augment the dataset. Their work is called \textup{Neural Augmentation}, which uses CycleGAN \cite{zhu2017unpaired} to create new stylised images in the training dataset. This method is finally tested on a 5-layer network with the MNIST dataset and delivers better performance than most traditional approaches. 
% \kg{Rephrase the next line. It is not clear what happens here.}
% They also explored to combine neural nets with the images classification network to build a new system, where the transformation neural nets can be trained with the classification loss so that they can identify the best augmentations for a given dataset.

% Another method called \textup{SamplePairing}, proposed by \cite{inoue2018data} also shows an improvement in the accuracy for image classification task, compared to traditional data augmentation strategies. Essentially, it tries to synthesise a new image from two images picked randomly from the training set. The synthesising method is straightforward but very effective in image classification tasks--- a new image is created based on the average value of each pixel of two images. It is also found that this technique can largely improve the accuracy when the training set is very small, so this data augmentation method is especially useful when the number of image examples available for training is limited, such as medical imaging tasks.

% \kg{In the following line, please explain how is speed related to limited training samples. It is not clear}
\section{Design and Implementation}\label{sec:des_n_imp}
\noindent In this section, we propose a modular design for using style transfer as data-augmentation technique. Figure \ref{3.1} summarises the modular architecture we followed for creating our data augmentation strategy. We choose the network designed in \cite{engstrom2016faststyletransfer} for our style transfer module. The reasons for this choice, architecture and implementation of the network are explained in subsection \ref{ssec:style_transfer_arch}. For testing the data augmentation technique we use the classification module. In our experiments we used the standard VGG16 from \cite{simonyan2014very} explained in section \ref{ssec:classification_arch}. In section \ref{ssec:datasets} we briefly explain the datasets used for evaluation of our strategy.

\subsection{Style Transfer Architecture}\label{ssec:style_transfer_arch}
\noindent For style transfer to be as viable as the other traditional data augmentation strategies (crop, flip \etc) we need a fast running solution. There has been successful style transfer solutions using CNNs but they are considerably slow \cite{gatys2016image}. To alleviate this problem we choose a generative architecture that only needs a forward pass to compute a stylised image \cite{engstrom2016faststyletransfer}. This network consists of a generative Style Transfer Network coupled with a Loss Network that computes a cumulative loss which can account for both style from the style image and content from the training image. In the following subsections (\ref{sssec:tranfer_net}, \ref{sssec:loss_net}) we will look at architectures of the Style Transfer Network and Loss Network. 

% \begin{figure*}[tb]
%   \centering
%   \includegraphics[scale=0.05]{style_module.png}
%   \caption{Style Transfer Module: Image transformation network is trained based on the loss computed by a pretrained loss network.}
%   \label{3.2}
% \end{figure*} 

\subsubsection{Transformation Network}\label{sssec:tranfer_net}
\noindent For the transformation network, we follow the state-of-the-art implementation from the work \cite{engstrom2016faststyletransfer} and \cite{resNets}, with changes to hyper parameters based on our experiments. 

%  For each residual block, we give an input tensor $x$. After a series of convolution layer, batch normalisation layer and ReLU layer, we get the output $F(x)$ from the residual block. Then the output tensor and the input tensor are summed into $H(x)$ and fed to following up layers. 
Five residual blocks are used in the style transformation network to avoid optimisation difficulty when the network gets deep \cite{he2016deep}. Other none residual convolutional layers are followed by Ulyanov's instance normalisation \cite{ulyanov1607instance} and ReLU layers. At the output layer, a scaled tanh is used to get an output image with pixels in the range from 0 to 255.

This network is trained coupled with the Loss Network (described in the following subsection) using stochastic gradient descent \cite{bottou2010large} to minimise a weighted combination of loss functions. We treat the overall loss as a linear combination of the content reconstruction loss and style reconstruction loss. The weights of two losses can be fine-tuned depending on the preference. By minimising the overall loss we can get a model well trained per style.

\subsubsection{Loss Network}\label{sssec:loss_net}
\noindent Since we already define the transformation network that can generate stylised images, we also need to create a loss network that is used to represent loss function to evaluate the generated images and use the loss to optimise the style transfer network based on stochastic gradient descent.

We use a deep convolutional neural network $\theta$ pretrained for image classification on imageNet to measure the texture and content differences between the generated image and the target image. Recent work has shown that deep convolutional neural networks can encode the texture information and high-level content information in the feature maps \cite{gatys2015texture}\cite{mahendran2015understanding}. Based on this finding, we define a content reconstruction loss $\ell_{c}^\theta$ and a style reconstruction loss $\ell_{style}^\theta$ in the loss network and use their weighted sum to measure the total difference between the stylised image and the image we want to get. For every style, we train the transformation network with the same pretrained loss network.

\textbf{Content Reconstruction Loss:} To achieve that, an image needs to be reconstructed from the image information encoded in the neural network, i.e., computing an approximate inverse from the feature map. Given an image $\vec{x}$, the image will go through the CNN model and be encoded in each layer by the filter responses to it. We use $F^{l}$ to store the feature maps in a layer $l$ where $F_{ij}^{l}$ is the feature map of the $i^{th}$ filter at position $j$ in layer $l$. Let $P^{l}$ be the feature maps for the content image in layer $l$, and we can update the pixels of image $x$ to minimise the loss $\mathcal{L}$ to make sure these two images have the similar feature maps in the network and thus have the similar semantic content:
\begin{equation}
  \mathcal{L}_{content} = \frac{1}{2}\sum_{i,j}(F_{ij}^{l} - P_{ij}^{l})^{2}
\end{equation}

\textbf{Style Reconstruction Loss:} We also want the generated images to have similar texture as the style target image, so we want to penalise the style differences with the style reconstruction loss. The feature maps in a few layers of a trained network are used for representing the texture by the correlations between them. Instead of using these feature maps directly, the correlations between the different channels of the feature maps are given by Gram matrix $G^{l}$, where the $G_{i,j}^{l}$ is the inner product between the vectorised $i^{th}$ feature map and $j^{th}$ feature map in layer $l$:

\begin{equation}
  G_{i,j}^{l} = \sum_{k}F_{ik}^{l}F_{jk}^{l}
\end{equation}

The original texture is passed through the CNNs and the Gram matrices $G^{l}$ on the feature responses of some layers are computed. We can pass a white noise image through the CNNs and compute the Gram matrix difference on every layer included in the texture model as the loss. If we use this loss to perform gradient on the white noise image and try to minimise the Gram matrix difference, we can find a new image that has the same style as the original image texture. The loss is computed by the mean-squared distance between the Gram matrix of two images. So let $A^{l}$ and $G^{l}$ be the Gram matrix of two images in layer $l$, the loss of that layer equal:

\begin{equation}
E_{l} = \frac{1}{4N^{2}_{l}M^{2}_{l}}\sum_{i,j}(G^{l}_{ij} - A^{l}_{ij})^{2}
\end{equation}

and the loss for all chosen layers:

\begin{equation}
  \mathcal{L}_{style} = \sum_{l=0}^{L} w_{l} E_{l}
\end{equation}

\textbf{Total Variation Regularization:} We also follow prior work \cite{gatys2016image} and make use of total variation regularizer $\mathcal{L}_{TV}$ to gain more spatial smoothness in the output image $\hat{y}$.

\textbf{Terms Balancing:} As we have above definitions, generating an image $\hat{y}$ can be seen as solving the optimising problem in the style transfer module in figure \ref{3.1}. We initialise the image with white noise, and the work \cite{gatys2016image} found that the initialisation has a minimal impact on the final results. $\lambda_{c}$, $\lambda_{s}$, and $\lambda_{TV}$ are hyperparameters that we can tune according to the monitoring of the results. To get the stylised image, we need to minimise a weighted combination of two loss functions and the regularization term:

\begin{equation}
\hat{y} = \operatorname*{argmin}_{y}\lambda_{c}\mathcal{L}_{content} + \lambda_{s}\mathcal{L}_{style} + \lambda_{TV}\mathcal{L}_{TV}
\end{equation}

\subsection{Image Classification}\label{ssec:classification_arch}
\noindent To evaluate the effectiveness of this design, we perform image classification tasks with the stylised images. In a image classification task, for each given image, the programs or algorithms need to produce the most reasonable object categories \cite{ILSVRC15}. The performance of the algorithm will be evaluated based on if the predicated category matches the ground truth label for the image. Since we will provide input images from multiple categories, the overall performance of an algorithm is the average score of overall test images.

Once we train the transformation network that can generate the stylised images, we apply it to the training dataset to create a larger dataset. The stylized images are saved on the disk with the ground truth categories. We then use them with their original images to train the neural networks to solve the image classification problems. In this research, the model we chose is VGGNet, which is a simple but effective model \cite{simonyan2014very}. Their team got the first place in the localisation and the second place in the classification task in ImageNet Challenge 2014. This model strictly used $3 \times 3$ filters with stride and pad of 1, along with $2 \times 2$ max-pooling layers with stride 2. 3 $3 \times 3$ convolutional layers back to back have an effective receptive field of $7 \times 7$. Compared with one $7 \times 7$ filter, $3 \times 3$ filter size can have the same effective receptive field with fewer parameters.

To fully understand the effectiveness of the style transfer and explore how useful style transfer can be compared and combined with other traditional data augmentation approaches for image classification problem, we need to experiment from multiple perspectives. The first experiments are to use traditional transformations alone. For each input image, we generate a set of duplicate images that are shifted, rotated, or flipped from the original image. Both the original image and duplicates are fed into the neural net to train the model. The classification performance will be measured on the validation dataset as the baseline to compare these augmentation strategies. The pipeline can be found in Figure \ref{3.1}. The second experiments are to apply the well-trained transformation network to augment the training dataset. For each input image, we select a style image from a subset of different styles from famous artists and use the transformation network to generate new images from the original image. We store the newly generated images on the disk, and both original and stylised images are fed to the image classification network. To explore if we can get better results, we go further to combine two approaches to get more images in training dataset .

\subsection{Dataset and Image Styles} \label{ssec:datasets}
\noindent The transformation network is trained on the COCO 2014 collection \cite{DBLP:journals/corr/LinMBHPRDZ14} containing more than 83k training images, which are enough to get the transformation network well trained. Since these images are used to feed the transformation network, we ignored the label information during training.

Two different datasets are used for images clasification tasks, caltech 101 \cite{fei2006one} and caltech 256 \cite{griffin2007caltech}. We keep training and testing on a fixed number of images and repeating the experiment with different augmented data and compare the results with others. The images are divided by a 70:30 split between training and validation for both datasets.

For the chosen styles, we try to select the images that look very different. At last, eight different images were chosen as the style input to train the transformation network. All styles can be found in the GitHub repo. \url{https://github.com/zhengxu001/neural-data-augmentation}.

% \textbf{Caltech101:} This dataset was collected by choosing 101 object categories from Google Images. There it contains 102 classes as there is one more background category. The total image number is 9144. For each class, there are 40 to 800 images, and images are roughly $300 \times 200$ pixels in size.

% \textbf{Caltech256:} This dataset has more than 30 thousands images. Compared with Caltech101, Caltech256 has several improvements. Firstly, the category number more than doubled. And then the minimum amount for each class is increased to 80. Since this dataset has much more images, it takes more time to train the classification network on this dataset.

% \textbf{Image Styles:} 
\section{\uppercase{Experiments}} \label{sec:experiments}
\noindent This section presents the evaluation of style transfer for data augmentation. We evaluate the results from multiple perspectives based on the classification Top-1 accuracy of VGGNet. The results of experiments on traditional augmentation are collected as the baseline. We then do experiments on every single style and some combined styles. We also combine the traditional methods with our style transfer method to verify their effectiveness and see if we can improve on previous methods.

\subsection{Traditional Image Augmentation}
\noindent We first used the pretrained VGGNet without any data augmentation and reached a classification accuracy of 83.34\% in one hour of training time. We then apply two different traditional image augmentation strategies, Flipping and Rotation, to train the model. Finally, we combine both strategies. Detailed results can be found in table \ref{5.1}. We found the model itself works the best without any augmentation. Using Flipping as data augmentation strategy gives very similar results, however the combination of \textup{Rotation} and \textup{Flipping} significantly reduces classification accuracy to 77\%. Adding \textup{Rotation} as data augmentation does not seem to help for classification, as is also shown in our following experiments.

\begin{table}[ht]
\centering
\def\arraystretch{1.2}
\begin{tabular}{ |c||c| }
 \hline
 \multicolumn{2}{|c|}{Traditional Image Augmentation} \\
 \hline
 Style Name& Result\\
 \hline
 None &0.8334\\
 Flipping&0.8305\\
 FlippingRotation&0.77\\
 \hline
\end{tabular}
\caption{Traditional Image Augmentation}
\label{5.1}
\end{table}

\subsection{Single Style}
\begin{figure}[!ht]
% \centering
\begin{subfigure}{0.145\textwidth}
  \includegraphics[width=\linewidth]{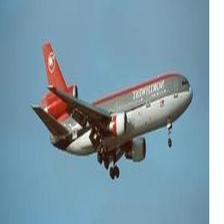}
  \caption{Original}
  \label{5.11}
\end{subfigure}%
\begin{subfigure}{0.145\textwidth}
  \includegraphics[width=\linewidth]{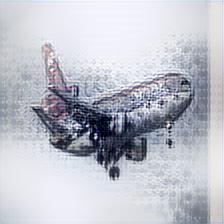}
  \caption{Snow}
  \label{5.12}
\end{subfigure}
\begin{subfigure}{0.145\textwidth}
  \includegraphics[width=\linewidth]{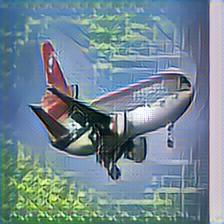}
  \caption{YourName}
  \label{5.13}
\end{subfigure}
\caption{Comparison between the original image and two stylized images. YourName style adds too many colours and shapes on the original image, which explains the bad performance in terms of the image classification accuracy.}
\label{5.20}
\end{figure}

\noindent We select eight different styles that look different from each other to train the transformation network. All styles can be found in the Appendix.  We feed each of the images in the training set to the eight fully trained transformation networks to generate eight stylised images. Both the original images and the stylised images are fed to VGGNet to train the network and the best validation accuracy from all epochs is recorded. The results for each style can be found in table \ref{5.2}. Compared with the traditional strategies, we can see that 7 out of 8 styles work better than the traditional strategies. It can also be seen that the \textup{Snow} style works the best, reaching an accuracy of 85.26\%, whereas YourName style only reaches 82.61\%. This is due to the addition of too much noise and colour to the original images for that particular style. In figure \ref{5.20} a comparison between the original image and two different stylised images is shown. As can be seen, YourName style adds too many colours and shapes on the original image, which explains the bad performance in terms of the image classification accuracy.

\begin{table}[ht]
\centering
\def\arraystretch{1.2}
\begin{tabular}{ |c||c|  }
 \hline
 \multicolumn{2}{|c|}{Single Style with VGG16 and Caltech 101} \\
 \hline
 Style Name&Result\\
 \hline
 Snow&\textbf{0.8526}\\
 RainPrincess&0.8494\\
 Scream&0.8490\\
 Wave&0.8468\\
 Sunflower&0.8461\\
 LAMuse&0.8436\\
 Udnie&0.8404\\
 Your Name&\textbf{0.8261}\\
 \hline
\end{tabular}
\caption{Single Style Transfer Augmentation}
\label{5.2}
\end{table}

\subsection{Combined Methods}
\noindent To evaluate the combination of two different styles, we take the original images and feed them to two different transformation networks and generate two stylized images for each input image. We then merge the stylized images and original images to compose the final training dataset. This gives us three times the number of images than the original dataset. We use this augmented data set to train the VGGNet model. We also try to combine the traditional augmentation methods with style transfer together and evaluate the performance. The results can be found in table \ref{5.3}.

\begin{table}[ht]
\centering
\def\arraystretch{1.2}
\begin{tabular}{ |c||c|c|}
 \hline
 \multicolumn{3}{|c|}{Combined Method} \\
 \hline
Traditional Method&Style&Result\\
 \hline
Flipping&None&0.8305\\
Flipping&Scream&0.8454\\
Flipping&Wave&\textbf{0.8486}\\
Flipping&ScreamWave&0.845\\
FlippingRotation&None&0.7521\\
FlippingRotation&Wave&0.7837\\
FlippingRotation&Scream&0.7862\\
FlippingRotation&ScreamWave&0.7942\\
\hline
\end{tabular}
\caption{Combined Style Transfer Augmentation}
\label{5.3}
\end{table}

We notice a very slight increase in performance of the combined style (ScreamWave) over the single styles. We further notice that adding Flipping as a data augmentation strategy degrades the performance, although to a lesser degree for the combined style.

% The original proportion for Wave style is $Content\_Weight:Style\_Weight = 9:100$. We increased the content weight to $Content\_Weight:Style\_Weight = 9:80$ 
\subsection{Content Weights Change}
\begin{figure}
\begin{subfigure}{.145\textwidth}
  \centering
  \includegraphics[width=\linewidth]{section5/originalImage.jpg}
  \caption{Original}
  \label{5.14}
\end{subfigure}%
\begin{subfigure}{.145\textwidth}
  \centering
  \includegraphics[width=\linewidth]{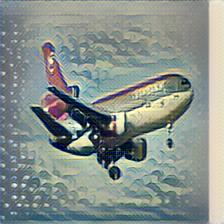}
  \caption{Wave2}
  \label{5.15}
\end{subfigure}
\begin{subfigure}{.145\textwidth}
  \centering
  \includegraphics[width=\linewidth]{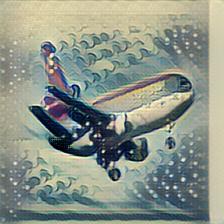}
  \caption{Wave}
  \label{5.16}
\end{subfigure}
\caption{Comparison between the original image (airplane) and stylized images with different content weights. (a) is the original image. Wave2 (b) has more content weight than Wave (c). Two stylized images look similar to each other.}
\label{5.17}
\end{figure}

\noindent In this experiment, we change the proportion of content weights and style weight in the transformation network to evaluate the impact on the performance. We increased the content weight to create a new style named Wave2. As can be seen from the table \ref{5.4}, no significant change can be observed for a change in content weights. As can be seen in figure \ref{5.17}, the images for the two content weights look very similar to each other, which explains the minimal impact in our experiments. In future work we would like to examine this effect further with more content weights.
\begin{table}[ht]
\centering
\def\arraystretch{1.2}
\begin{tabular}{ |c||c|c| }
 \hline
 \multicolumn{3}{|c|}{Different Content Weights} \\
 \hline
Traditional Method&Style Name&Result\\
 \hline
None&Wave&0.8468\\
None&Wave2&0.8417\\
\hline
\end{tabular}
\caption{Wave2 gets more content weight than Wave}
\label{5.4}
\end{table}

\subsection{VGG19}
\noindent To evaluate the generalisation of our approach over network architectures, we used VGG19 as a classification network and duplicated our experiments. The results can be found in table \ref{5.24}.

\begin{table}[ht]
\centering
\resizebox{0.35\textwidth}{!}{
\def\arraystretch{1.2}
\begin{tabular}{ |c||c|c| }
\hline
\multicolumn{3}{|c|}{Experiments on VGG19} \\
\hline
Traditional Method& Style Name&Result\\
\hline
None&None&0.8450\\
Flipping&None&0.8367\\
None&Wave&0.8425\\
Flipping&Wave&\textbf{0.8581}\\
None&Scream&0.8446\\
Flipping&Scream&0.8465\\
% FlippingRotation&ScreamWave&0.8029\\
\hline
% 256
None&None&0.62\\
Flipping&None&0.6666\\
None&Scream&0.6313\\
Flipping&Scream&\textbf{0.6728}\\
None&Wave&0.6272\\
Flipping&Wave&0.6632\\
% Flipping Rotation&None&0.5853\\
\hline
\end{tabular}}
\caption{Experiments on VGG19 with Caltech datasets}
\label{5.24}
\end{table}

The baseline classification accuracy for Caltech 101 is 84.5\%. Based on this number, there are some interesting findings in line with the VGG16. Using \textup{Wave} or \textup{Flipping} itself does not improve the performance of VGG19, but if we combine \textup{Flipping} with \textup{Wave} we can get a considerable improvement, with accuracy reaching 85.81\%.

Similar results can be seen for our experiments on Caltech 256. The use of \textup{Flipping} as data augmentation gives an accuracy of 66.66\%, while the use of \textup{Scream} gives at 63.13\%. However, if we combine the two approaches, we can get an accuracy of 67.28\%. The combination between \textup{Flipping} and \textup{Wave} gives 66.32\% accuracy which is higher than for \textup{Wave} alone. 

% Mirroring our earlier results, it can be seen that adding \textup{Rotation} as data augmentation decreases the performance over all combinations.

The experiments performed on VGG19 show that the style transfer is still an effective data augmentation method, which can be combined with the traditional approaches to further improve the performance.
\section{\uppercase{Conclusions}}
\noindent In this paper, we proposed a novel data augmentation approach based on neural style transfer algorithm. From our experiments, we observe that this approach is an effective way to increase the performance of CNNs in image classification tasks. The accuracy for VGG16 is increased to 85.26\% from 83.34\% while for VGG19 the accuracy increased to 85.81\% from 84.50\%. We also found that we can combine this new approach with traditional methods like flipping, rotation \etc to boost the performance. 

We tested our method only for image classification task. As a future work, it would be interesting to try it out for other computer vision tasks such as segmentation, detection \etc. The set of styles is also limited. Even though we tried to select images of different styles, we did not classify the images according to their category. A more elaborate set of styles might train more robust models. Another limitation of this approach is the speed of training, which is quite slow. However, once trained, the inference can still be fast as it does not involve augmentation. 

% \textbf{Styles Category:} \kg{Do we need to be so specific? It should be fine to point towards some high level problems. You can rephrase this para into a single line and merge it with the last para}  Even though we tried to select images of different styles, we did not classify the images according to their category. We do not know if an image is Abstract, Classical or Expressionism. Maybe in the future, we can choose several images that belong to one category to test if the category has an impact on the image classification performance, or if different images in the same category have different effectiveness.

Our approach is independent of the architecture. Better and faster models for style transfer will enable more diverse and robust augmentation for a CNN. We hope that the proposed approach will be useful for computer vision research in future.

\section*{Acknowledgments}
This publication has emanated from research conducted with the financial support of Science Foundation Ireland (SFI) under the Grant Number 15/RP/2776.
\bibliographystyle{apalike}
{\small\bibliography{ms}}
\end{document}